# Machine Learning Classification of Alzheimer's Disease Stages Using Cerebrospinal Fluid Biomarkers Alone

Vivek Kumar Tiwari[1], Premananda Indic[1], Shawana Tabassum[1#]

[1]Department of Electrical Engineering, University of Texas, Tyler; Texas, U.S.A.

[#]Corresponding Author: Shawana Tabassum: stabassum@uttyler.edu

## Abstract:

Early diagnosis of Alzheimer's disease is a challenge because the existing methodologies do not identify the patients in their preclinical stage, which can last up to a decade prior to the onset of clinical symptoms. Several research studies demonstrate the potential of cerebrospinal fluid biomarkers, amyloid beta 1-42, T-tau, and P-tau, in early diagnosis of Alzheimer's disease stages. In this work, we used machine learning models to classify different stages of Alzheimer's disease based on the cerebrospinal fluid biomarker levels alone. An electronic health record of patients from the National Alzheimer's Coordinating Centre database was analyzed and the patients were subdivided based on mini-mental state scores and clinical dementia ratings. Statistical and correlation analyses were performed to identify significant differences between the Alzheimer's stages. Afterward, machine learning classifiers including K-Nearest Neighbors, Ensemble Boosted Tree, Ensemble Bagged Tree, Support Vector Machine, Logistic Regression, and Naïve Bayes classifiers were employed to classify the Alzheimer's disease stages. The results demonstrate that Ensemble Boosted Tree (84.4%) and Logistic Regression (73.4%) provide the highest accuracy for binary classification, while Ensemble Bagged Tree (75.4%) demonstrates better accuracy for multiclassification. The findings from this research are expected to help clinicians in making an informed decision regarding the early diagnosis of Alzheimer's from the cerebrospinal fluid biomarkers alone, monitoring of the disease progression, and implementation of appropriate intervention measures.

# Introduction

Alzheimer's disease (AD) is one of the most prevalent neurodegenerative diseases characterized by progressive cognitive decline. This cognitive deterioration also poses a significant burden on the caregivers. Despite the advances in disease interventions, the gap in knowledge for diagnosis and effective management of AD is threefold. First, biochemicals drive most physiological signals, yet the search for molecular biomarkers for precise and early detection of the pathology of AD remains a challenge. Biomarkers are biological molecules (for instance proteins) found in body fluids and indicate the stage of disease as well as the response to treatment. The well-established biomarkers of dementia include amyloid-beta peptide 1-42 (Aβ1-42), total tau (T-tau), and tau phosphorylated at threonine (P-tau) [1]. However, much is yet to be understood regarding the role of these molecular biomarkers in early diagnosis and decision-making for AD. Second, AD is often not diagnosed at an early stage because of the lack of a screening tool that can monitor/track the pathophysiological processes at the bedside. It is evident from positron emission tomography (PET) imaging studies that the underlying progressive pathology precedes the symptomatic onset of AD by one or two decades [2]. The underlying disease process likely begins years before AD manifests clinically [3]. As a result, significant irreversible neurological damage occurs by the time dementia is diagnosed, thereby reinforcing the need for point-of-care diagnosis for monitoring the pathogenesis of AD and identifying the patients in predementia stages. Third, there is a lack of a classification framework for analyzing the time-series biomarker levels and making an informed decision about timely diagnosis and clinical interventions.

The mild impairment of episodic memory is typically one of the first indications of patients with early-stage AD. These people may meet the requirements for mild cognitive impairment (MCI), but because their daily life activities are unaffected and their global cognitive functioning is intact, they are not yet demented [4, 5]. In addition, staging plays a crucial role in selecting the right pharmacologic therapy options that have been approved for varying levels of illness severity. The clinical dementia rating (CDR), initially published in 1982 [6], is a gold standard global rating scale for staging dementia patients. The CDR classifications are zero (no dementia), 0.5 (questionable dementia), one (mild dementia), two (moderate dementia), and three (severe dementia). Although CDR is highly valid and reliable, the score relies on a comprehensive structured interview with both the patient and the physician [7, 8]. In addition, the process becomes complicated if a trustworthy and knowledgeable caregiver is unavailable. As a result, CDR has little utility in everyday practice [9]. In contrast to CDR, simple and brief assessment techniques such as the mini mental state examination (MMSE) scores are more appropriate at the primary and secondary levels of care [10]. The MMSE score (ranging from 0 to 30) is calculated based on a series of questions designed to assess the cognitive skills of the patient. A score of 26 or higher represents normal cognition, scores between 20 and 26 indicate mild dementia, scores between 10 and 20 indicate moderate dementia, and scores less than 9 indicate severe dementia. Numerous studies have demonstrated the reliability and validity of the MMSE score [11].

In this work, we have used machine learning models for the classification of AD stages based on the cerebrospinal fluid (CSF) biomarkers alone. CSF is a colorless, watery fluid that is found in the brain and spinal cord. We have used MMSE and CDR scores to subdivide AD patients into stages such as normal cognition (NC), mild cognitive impairment (MCI), and severe dementia (SD). Afterward, we performed binary (NC vs any other dementia) and multiclassification (NC vs MCI vs SD) operations to classify AD stages using six different machine-learning classifiers. The findings show that when the Alzheimer's stages were split based on CDR, Ensemble Boosted Tree provided the highest accuracy of 84.4% in terms of binary classification, while Ensemble Bagged

Tree provided an accuracy of 75.4% for multiclassification. However, when the Alzheimer's stages were sub-divided using the MMSE score, both Boosted Tree and Logistic Regression models showed a better accuracy of 73.3% and 73.4%, respectively, for binary classification, and the Ensemble Bagged Tree classifier exhibited a higher accuracy of 56.8% for multi-classification.

## Significance of this Research

AD diagnosis has been extensively researched in the past decades. The majority of the current research is either limited to identifying AD patients [12] or projecting the progression of the disease over a predetermined amount of time [13]. These are presented as binary classification (e.g., NC vs. AD [14] and MCI vs. SD [15]) or multiclass classification tasks (e.g., NC vs. MCI vs. SD) [16]. Only a few research studies [17, 18] approached the problem as a regression challenge, and the models used in these studies fit logistic or polynomial functions to the longitudinal dynamics of each biomarker individually. Additionally, some studies [19, 20–24] classified the patients or predicted AD progression using only magnetic resonance imaging (MRI) scans. Rallabandi et al. [19] employed a support vector machine (SVM) on multiple cortical thickness measures from structural MRI scans to classify NC, early MCI, late MCI, and AD groups, with an overall accuracy of 75%. Although the proposed method led to a higher false positive rate for classifying late MCI and AD, the model showed promise in distinguishing early and late MCI groups. In order to predict the Progressive MCI (pMCI) class, Jin et al. [25] obtained 16 characteristics from MRI scans and combined them with the MMSE score. Using a boosted tree ensemble classifier, they were able to predict AD with an accuracy of 56.25%. These models sought to identify the patient's precise stage, such as NC, MCI, or AD. However, the precision of these results is insufficient to enable medical decision-making because they rely on a single modality. Another research project called Alzheimer's Biomarkers in Daily Practice (ABIDE) used the Amsterdam Dementia Cohorts (longitudinal cohort, tertiary referral center) of 525 individuals [26]. Every patient's initial appointment took place in a clinic between September 1, 1997, and August 31, 2014. In this study, Cox regression (or proportional hazards regression) analysis was performed to develop prognostic models for analyzing the progression of Mild Cognitive AD phases. MRI biomarkers (hippocampal volume and normalized whole-brain volume), cerebrospinal fluid (CSF) biomarkers (amyloid-1-42, tau), patient characteristics such as age and gender, and MMSE scores were used as input to the prognostic models. However, this study did not classify all stages of Alzheimer's disease. Moreover, the accuracy of the prognostic models relied on MRI scans, which are expensive to acquire and require specialized expertise to decode. In this regard, multi-stage classification of Alzheimer's disease stages based on only CSF biomarkers would entail a cost-effective and early diagnostic solution.

Recent research on CSF biomarkers has revealed the potential to detect AD at its early stages, in MCI patients [27]. Initial research revealed that T-tau and Aβ42 together had a good predictive potential to detect early-stage AD in MCI cases [27]. According to recent studies with prolonged periods of clinical follow-up, the combination of all three CSF biomarkers (T-tau, P-tau, and Aβ1-42) may have a predictive value as high as 95% to distinguish MCI cases with progression toward AD from stable MCI cases and MCI cases with other types of underlying pathology [28]. Several large multi-center studies have confirmed that these CSF biomarkers have a strong predictive value for detecting early-stage AD [29-31]. The significant contributions of this research are the following:

- Classifying different stages of AD based on the CSF biomarkers (Aβ1-42, P-tau, T-tau, and the ratio of Aβ1-42/P-tau) alone.

- Machine learning models including Logistic Regression, Boosted Tree, Bagged Tree, Naïve Bayes, Support Vector Machine, and K-Nearest Neighbors were used for binary and multi-classification of AD stages.
- Patients were sub-divided based on both MMSE and CDR scores and a comparative analysis was performed.

• Demonstrating the higher accuracy of CDR in classifying the AD stages as compared to the MMSE score.

## Methodology

**Description of Data:** An electronic health record of AD patients collected from the National Alzheimer's Coordinating Center (NACC) database was analyzed in this work [32]. We analyzed two separate datasets, one containing the CSF biomarker levels and the other containing the MMSE and CDR scores. Firstly, the data merging technique was performed in Python to join the MMSE scores with corresponding CSF biomarker levels and then CDR scores with CSF biomarker levels. This step allowed dropping the patients with missing scores or biomarker levels. All patients underwent lumbar puncture for collecting cerebrospinal fluid (CSF) and the reported levels of $A\beta 1$-42, P-tau, and T-tau were in pg/ml.

**Statistical Analysis:** Table I(A) shows the patient age, MMSE score, and CSF biomarker levels in terms of mean ($\pm$standard error). Compared to the individuals with NC, the AD patients (i.e., SD, MCI, and moderate dementia (MOD)) had considerably lower levels of $A\beta 1$-42, higher levels of T-tau, and a lower $A\beta 1$-42/P-tau ratio. Pearson correlation coefficients (r) were computed to investigate the association between MMSE scores (that classify NC, MCI, MOD, and SD) and

CSF biomarker levels (Aβ1-42, P-tau, and T-tau, and Aβ1-42/P-tau), as listed in Table I(B). Additionally, we used a one-way Analysis of Variance (ANOVA) to determine whether there were statistically significant differences between individual biomarker levels and the MMSE score. A p-value of less than 0.05 was considered statistically significant. Our findings indicate the following:  1) there was no statistically significant difference between subjects of different ages, hence the results are not included in Table I(B), 2) when the total number of patients was considered, all the biomarkers showed some association with the cognitive function, 3) the Aβ1-42 levels had a statistically significant negative correlation with the MMSE score in the SD group, but a statistically significant positive correlation with the MMSE score in the NC group, 4) T-tau levels had a statistically significant positive correlation with the MMSE score in the SD group, but a statistically significant negative correlation with the MMSE score in the NC group, and 5) the P-tau levels in the MOD and MCI groups had a statistically significant positive association with the MMSE score, but a statistically significant negative correlation with the MMSE score in the NC group.

TABLE I
(A) Comparison Among the SD, MOD, MCI, and NC Groups with Respect to Mean and Standard Error.
(B) Correlation (Through Pearson Correlation Coefficient, r) and Significant Difference (p Values)
Between the CSF Biomarker Levels and the MMSE Score

| Variables | SD (n=20) | MOD (n=101) | MCI (n=152) | NC (n=167) |
|---|---|---|---|---|
| Age | 68.41 | 69.43 | 71.23 | 66.85 |
| (years) | (2.51) | (0.78) | (0.75) | (0.44) |
| MMSE | 5.580 | 16.35 | 22.72 | 28.47 |
| | (0.88) | (0.24) | (0.09) | (0.06) |
| $A\beta_{1-42}$ | 232.88 | 193.72 | 226.43 | 369.7 |
| (pg/ml) | (38.28) | (8.57) | (2.51) | (10.00) |
| T-tau | 321.99 | 195.5 | 261.3 | 150.8 |
| (pg/ml) | (133.25) | (24.64) | (42.06) | (9.73) |
| P-tau | 42.5 | 53.13 | 45.24 | 37.97 |
| (pg/ml) | (4.92) | (2.75) | (3.17) | (1.06) |
| $A\beta_{1-42}$/P-tau | 7.05 | 5.55 | 7.71 | 12.31 |
| | (1.82) | (0.47) | (0.59) | (0.33) |

**(A)**

| | Total Number of patients | SD | MOD | MCI | NC |
|---|---|---|---|---|---|
| $A\beta_{1-42}$ (pg/ml) | r=0.33 p=0.001 | r=-0.64 p=0.005 | r=0.15 p=0.06 | r=-0.17 p=0.83 | r=0.26 p=0.006 |
| T-tau (pg/ml) | r=-0.11 p<0.001 | r=0.45 p=0.006 | r=0.17 p=0.03 | r=-0.04 p=0.63 | r= -0.12 p=0.0025 |
| P-tau (pg/ml) | r=-0.17 p=0.001 | r=0.04 p=0.87 | r=0.19 p=0.01 | r=0.03 p=0.01 | r=-0.20 p=0.004 |
| $A\beta_{1-42}$/P-tau | r=0.348 p<0.001 | r=-0.108 p=0.679 | r=-0.019 p=0.815 | r=0.021 p=0.80 | r=0.298 p<0.001 |

**(B)**

**Machine Learning-based Classification Models:** Followed by statistical analysis, we used machine learning models to classify different stages of dementia based on CSF biomarker levels. Machine learning models including Logistic Regression, Boosted Tree, Bagged Tree, Naïve Bayes, Support Vector Machine, and K-Nearest Neighbors (KNN) were used for binary classification of Alzheimer's disease stages. In contrast, Boosted Tree and Bagged Tree were used for multi-level classification. By examining the confusion matrices of Logistic Regression, Naive Bayes, Support Vector Machine, and K-Nearest Neighbors (KNN), we found that although these machine learning classifiers had acceptable classification accuracy, they were unable to track data in SD patients. Hence, we only used Bagged Tree and Boosted Tree classifiers for the multi-level classification of AD stages. Previous research in the literature shows the promising performance of these models in classification problems [15]. We used MATLAB to apply the machine learning classification models to the NACC database. Subsequently, we computed the Receiver Operating Characteristic (ROC) curve, confusion matrix, true positive rate, and false positive rate, to evaluate the performance of the machine learning models. In addition, a comparative analysis is provided on the performance of machine learning models when patients were subdivided based on CDR ratings versus MMSE scores.

### a) Binary Classification of Alzheimer's Disease

This section outlines the outcomes from the binary classification of NC individuals from other dementia stages. The Aβ1-42, P-tau, T-tau, and Aβ1-42/P-ratio were used as the input parameters to machine learning classifiers. Figure 1a depicts the inputs and outputs of the binary classifier that classifies AD patients from NC individuals.

First, we selected those patients from the NACC database who were staged based on the Mini Mental Examination (MMSE) scores. The total number of patients was 696 after merging the MMSE scores with corresponding CSF biomarker levels. As indicated in Figure 1, 273 patients were diagnosed with a certain stage of AD, while 423 patients belonged to the NC category. In contrast, when Clinical Dementia Rating (CDR) scale was used to subdivide the patients, there were a total of 558 patients, wherein 227 were diagnosed with AD and 331 belonged to the NC category. In both cases (MMSE versus CDR), a 5-percent cross-validation method was applied before the data were fed into the machine learning classifiers.

Our results showed that logistic regression provided the highest accuracy in classifying MMSE-staged patients into AD and NC categories. Logistic regression, also referred to as logit regression, log-linear classifier, or maximum-entropy classifier, is a linear classification model. The (binary) logistic model is a statistical model used to predict the likelihood of one event occurring out of two options. The model achieves this by creating a linear combination of independent variables, also known as "predictors" [33]. In this study, we have used logistic regression as a classifier for binary as well as multi-classification.

The confusion matrix is shown in Figure 2a, where 172 patients were categorized as AD patients and 339 individuals were accurately identified as NC by the logistic regression classification model. Figure 2b depicts the ROC plot with the area under the curve (AUC) found to be 0.78. The performance comparison of the machine learning models that we utilized for binary classification is shown in Table II (with patients subdivided by the MMSE score). The following performance metrics were compared: accuracy and True Positive Rate (TPR). It can be observed that logistic regression demonstrates a higher accuracy (73.4%) as well as higher TPRs (80.1% for NC and 63.0% for AD).

In contrast, the boosted tree classifier showed higher accuracy (84.4%) as compared to other machine learning classifiers for CDR-staged patients. Ensemble tree-based models can successfully learn from data having a variety of attributes. Figure 2c shows the confusion matrix for boosted tree classifier, which accurately identified 296 NC individuals and 175 AD patients. Figure 2d shows the corresponding ROC curve and the resulting AUC (=0.88). Table III shows the performance comparison of the machine learning models, used for binary classification (with patients subdivided by the CDR rating). It is evident that Boosted Tree shows a higher accuracy (84.4%) along with reasonably high TPRs (89.4% for NC and 77.1% for AD).

The accuracy, TPR, False Negative Rate (FNR), Positive Predictive Value (PPV), and False Discovery Rate (FDR) for binary classification of AD stages are shown in Figures 3 and 4. The definition of the classification models used in this work, along with the terms TPR, FNR, PPV, and FDR are provided in the Appendix.

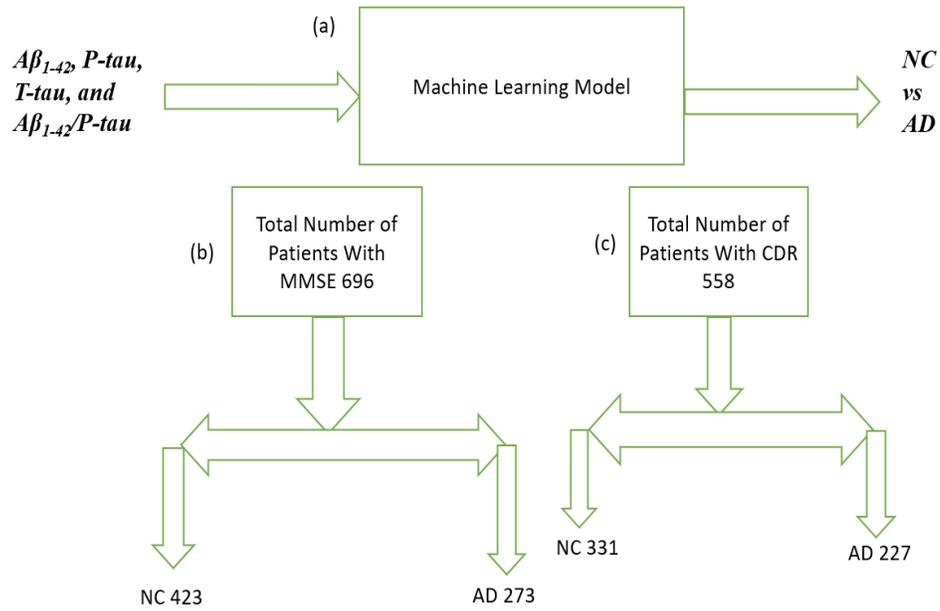

FIGURE 1. (a) Input and output of the binary classifier. Patients were subdivided into normal cognition (NC) and Alzheimer's disease (AD) groups based on (b) MMSE and (c) CDR scores.

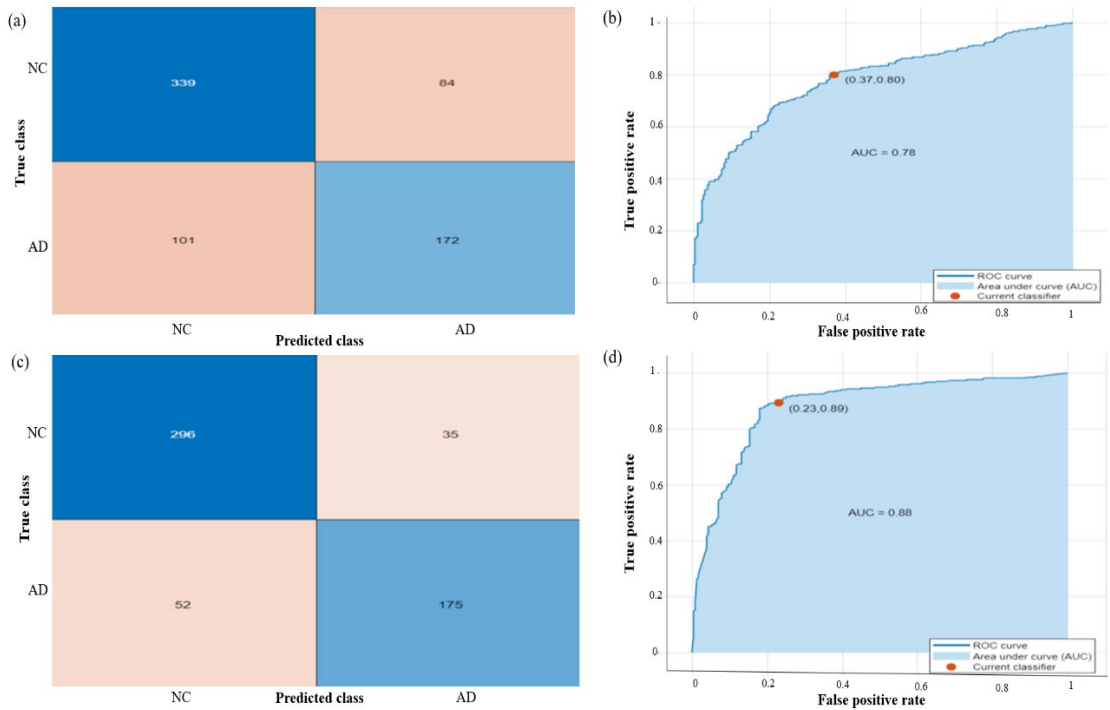

FIGURE 2. (a) Confusion matrix and (b) Receiver operating characteristic (ROC) curve for binary classification of MMSE subdivided patients using Logistic regression. (c) Confusion matrix and (d) ROC curve for binary classification of CDR subdivided patients using Boosted Tree.

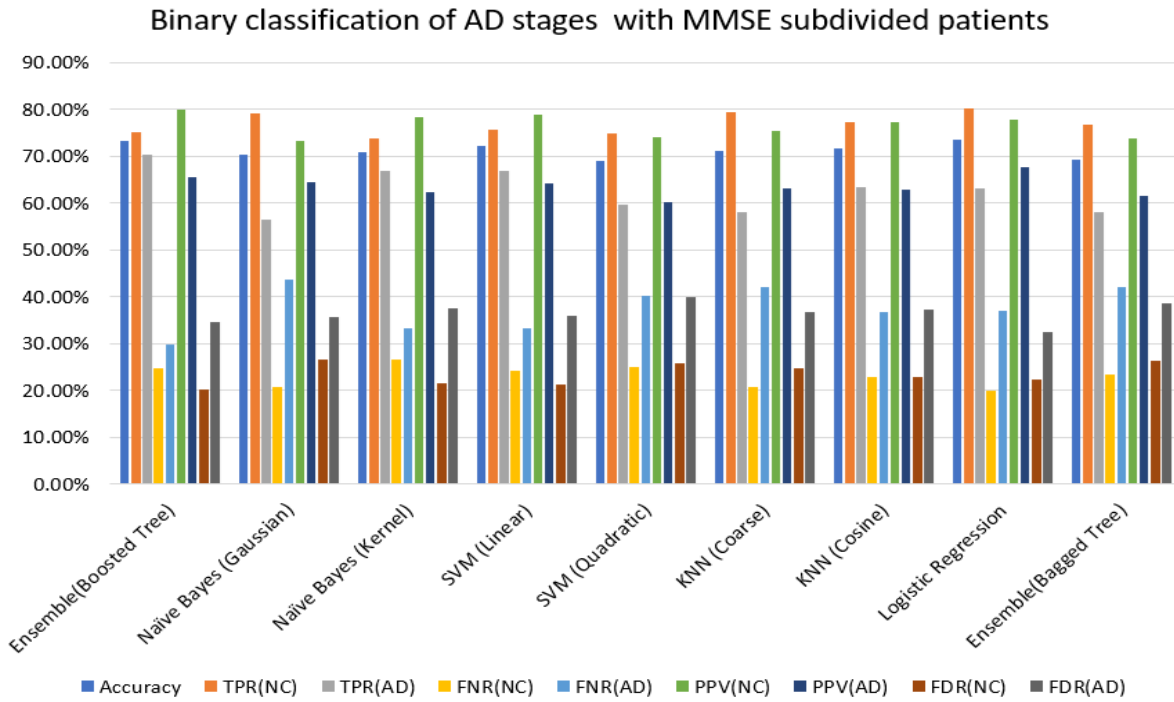

FIGURE 3. Performance comparison of binary classification models with MMSE subdivided patients.

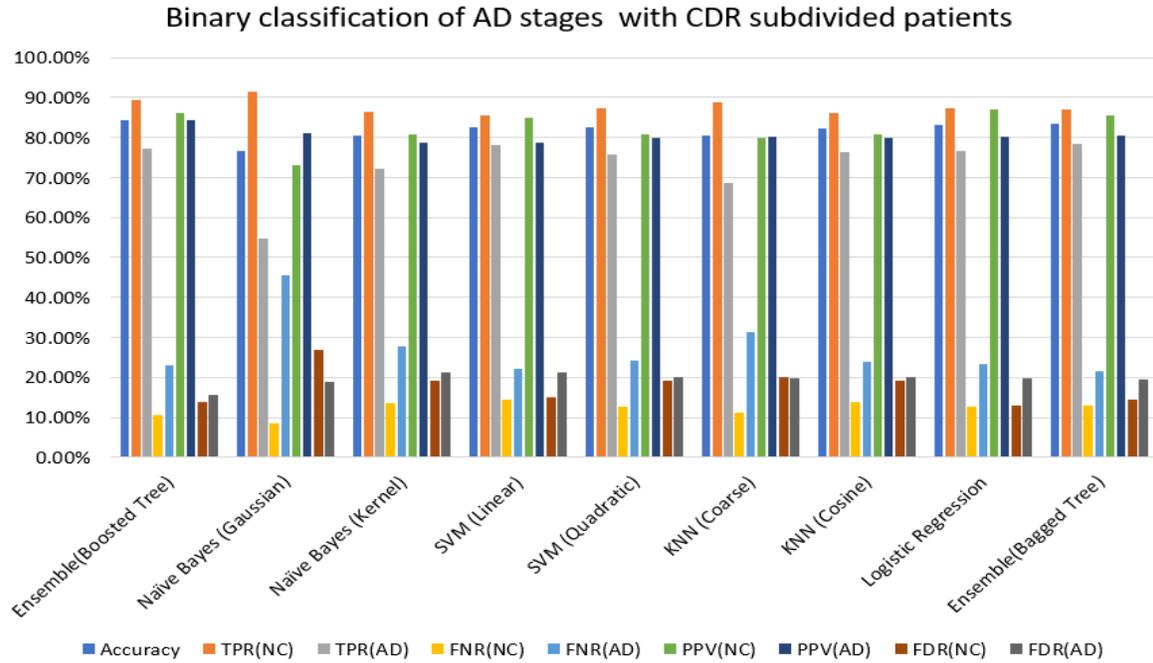

FIGURE 4. Performance comparison of binary classification models with CDR subdivided patients.

TABLE II
PERFORMANCE COMPARISON OF MACHINE LEARNING MODELS FOR BINARY CLASSIFICATION OF MMSE-STAGED
PATIENTS

| Machine Learning Models | ACCURACY | TPR (NC) | TPR (AD) |
|---|---|---|---|
| Ensemble (Boosted Tree) | 73.3% | 75.2% | 70.3% |
| Naïve Bayes (Gaussian) | 70.3% | 79.2% | 56.4% |
| Naïve Bayes (Kernel) | 70.7% | 73.7% | 66.7% |
| SVM (Linear) | 72.1% | 75.7% | 66.7% |
| SVM (Quadratic) | 69.0% | 74.9% | 59.7% |
| KNN (Coarse) | 71.% | 79.4% | 57.9% |
| KNN (Cosine) | 71.7% | 77.1% | 63.4% |
| Logistic Regression | 73.4% | 80.1% | 63.0% |
| Ensemble (Bagged Tree) | 69.3% | 76.6% | 57.9% |

| Machine Learning Models | ACCURACY | TPR (NC) | TPR (AD) |
|---|---|---|---|
| Ensemble (Boosted Tree) | 84.4% | 89.4% | 77.1% |
| Naïve Bayes (Gaussian) | 76.5% | 91.5% | 54.6% |
| Naïve Bayes (Kernel) | 80.6% | 86.4% | 72.2% |
| SVM (Linear) | 82.4% | 85.5% | 78.0% |
| SVM (Quadratic) | 82.6% | 87.3% | 75.8% |
| KNN (Coarse) | 80.6% | 88.8% | 68.7% |
| KNN (Cosine) | 82.1% | 86.1% | 76.2% |
| Logistic Regression | 83.0% | 87.3% | 76.7% |
| Ensemble (Bagged Tree) | 83.5% | 87.0% | 78.4% |

**b)** **Multi-Classification of Alzheimer's Disease with an Imbalanced Dataset**

For multi-classification, the patients were divided into three classes, NC, MCI, and SD, as illustrated in Figure 5. We used RUS Boosted Tree and Bagged Tree classifiers for multi-classification. Merging the MMSE score with CSF biomarker levels resulted in a total of 696 patients, with 423 individuals identified with NC, 152 with MCI, and 121 with SD, as shown in Figure 5b. In contrast, joining the CDR rating with CSF biomarker levels resulted in a total of 558 patients, with 331 people in the NC stage, 191 in the MCI stage, and 36 in the SD stage, as listed in Figure 5c.

The Ensemble Bagged Tree shows better accuracy (56.8%) in comparison to the RUS Boosted Tree model for MMSE-staged patients. For CDR subdivided classes, the bagged tree classifier had better accuracy (75.4%) in comparison to RUS Boosted Classifier. However, when the TPR and FNR values are considered for the SD group, RUS boosted is found to provide higher TPR and lower FNR values in comparison to the bagged tree classifier. Hence, overall, RUS Boosted Tree demonstrates better performance than Ensemble Bagged Tree. The confusion matrix of the RUS Boosted Tree model for MMSE-staged patients is shown in Figure 6a. It can be observed that 268 patients were accurately classified as NC, 34 patients were classified as MCI, and 70 as SD for MMSE-staged patients. Figure 6b shows the ROC curve for the RUS Boosted Tree model in which Area Under the Curve (AUC) was found to be 0.76. Figure 6c shows the confusion matrix of the RUS Boosted tree classifier that accurately classifies 263 patients as NC, 81 patients as MCI, and 14 as SD for CDR-staged patients. Figure 6d shows the ROC curve for the RUS Boosted Tree model where AUC was found to be 0.87.

Table IV and Table V show the comparison of RUS Boosted and Bagged Tree models that we used for the multi-classification of MMSE and CDR subdivided patients. The tables again demonstrate that RUS Boosted showed better performance for patient staging with both MMSE and CDR. Figures 7 and 8 show the comparison of the classifiers with respect to accuracy, TPR, FNR, PPV, and FDR. The Bagged Tree classifier showed better accuracy in both MMSE and CDR-subdivided patients, while RUS Boosted provided higher TPR and lower FNR in the SD group. Therefore, it can be concluded that the performance of the RUS Boosted model is better compared to Bagged Tree.

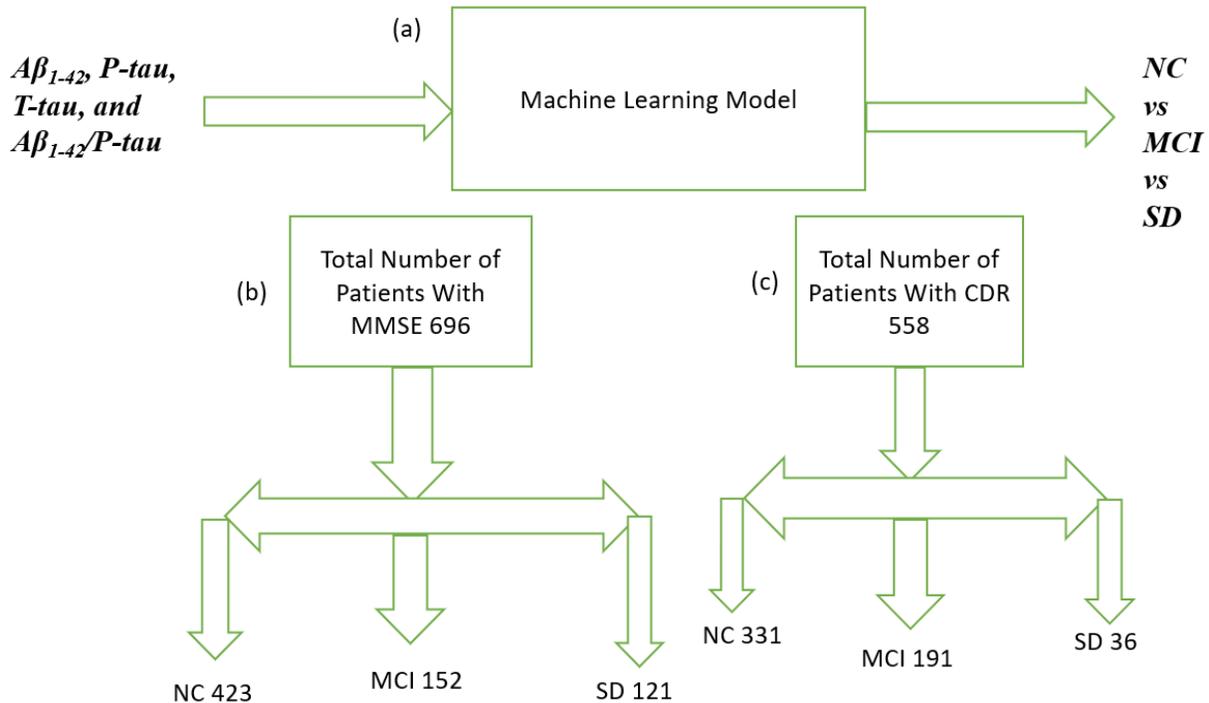

**FIGURE 5.** (a) Input and output of the multiclass classifier. Patients were subdivided into normal cognition (NC), mild cognitive impairment (MCI), and severe dementia (SD) groups based on (b) MMSE and (c) CDR scores.

### c) **Multi-Classification of Alzheimer's Disease with a Balanced Dataset**

In the imbalanced dataset, the NC class was over-represented. Hence, in order to prevent the models from becoming biased towards one class, we adopted the approach of balancing the dataset for both MMSE and CDR-subdivided patients. We employed the undersampling balancing strategy by removing samples from the over-represented class (in this case, NC) until all the groups had an equal distribution of data. Balancing was done to investigate whether the machine learning classifiers performed better in classifying the AD stages. For MMSE-staged patients, there were 121 patients in each of the NC, MCI, and SD categories. Likewise, for CDR-staged patients, there were 36 patients in each patient group.

The Ensemble Bagged Tree shows better accuracy (46.3%) in comparison to the Ensemble Boosted Tree model for MMSE-staged patients. For CDR subdivided classes, the bagged tree

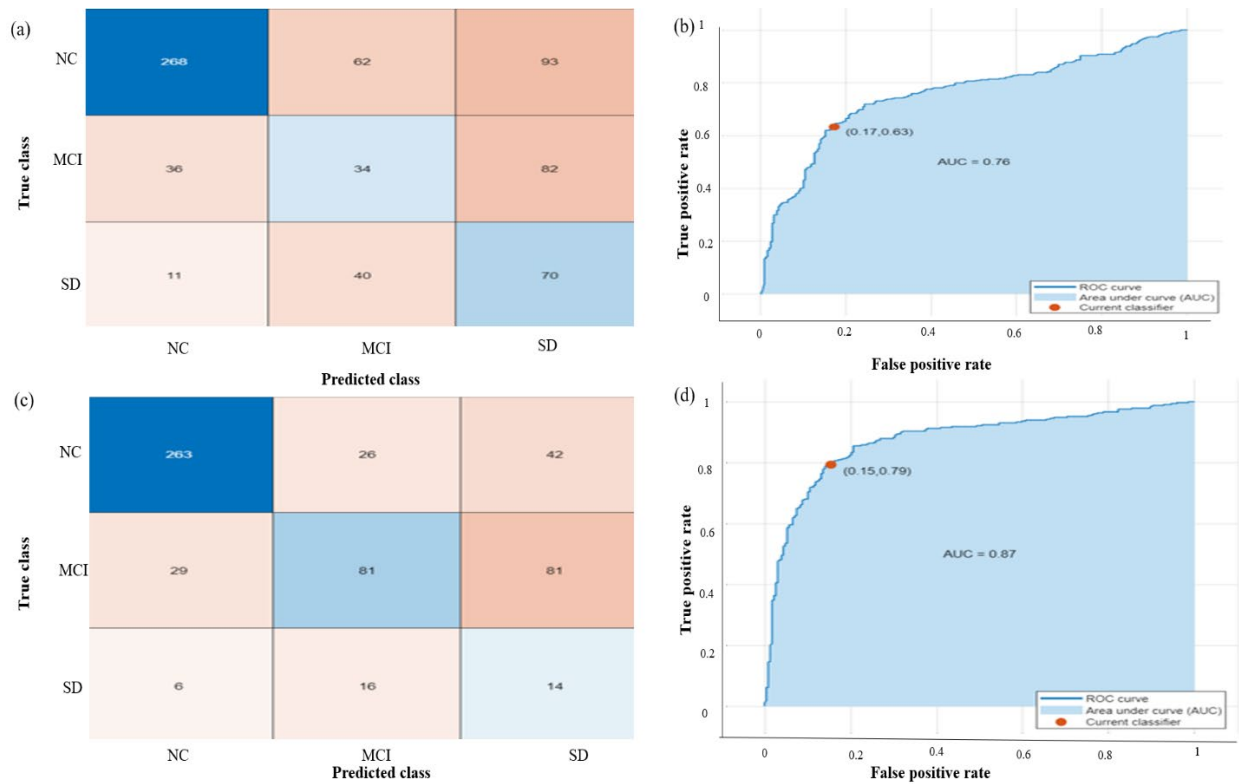

FIGURE 6. Machine learning classification applied to the imbalanced dataset. (a) Confusion matrix and (b) Receiver operating characteristic (ROC) curve for multiclass classification of MMSE subdivided patients using Ensemble Boosted Tree. (c) Confusion matrix and (d) ROC curve for multiclass classification of CDR subdivided patients using Boosted Tree.

TABLE IV
COMPARISON OF CLASSIFICATION MODELS FOR MULTI CLASSIFICATION OF MMSE-STAGED PATIENTS.

| Machine Learning Models | Accuracy | TPR (NC) | TPR (MCI) | TPR (SD) |
|---|---|---|---|---|
| Ensemble (RUS Boosted Tree) | 53.4% | 63.4% | 22.4% | 57.9% |
| Ensemble (Bagged Tree) | 56.8% | 77.5% | 27.6% | 20.7% |

TABLE V

COMPARISON OF CLASSIFICATION MODELS FOR MULTI CLASSIFICATION OF CDR-STAGED PATIENTS.

| Machine Learning Models | Accuracy | TPR (NC) | TPR (MCI) | TPR (SD) |
|---|---|---|---|---|
| Ensemble (RUS Boosted Tree) | 64.2% | 79.5% | 42.4% | 38.9% |
| Ensemble (Bagged Tree) | 75.4% | 88.2% | 67.0% | 2.8% |

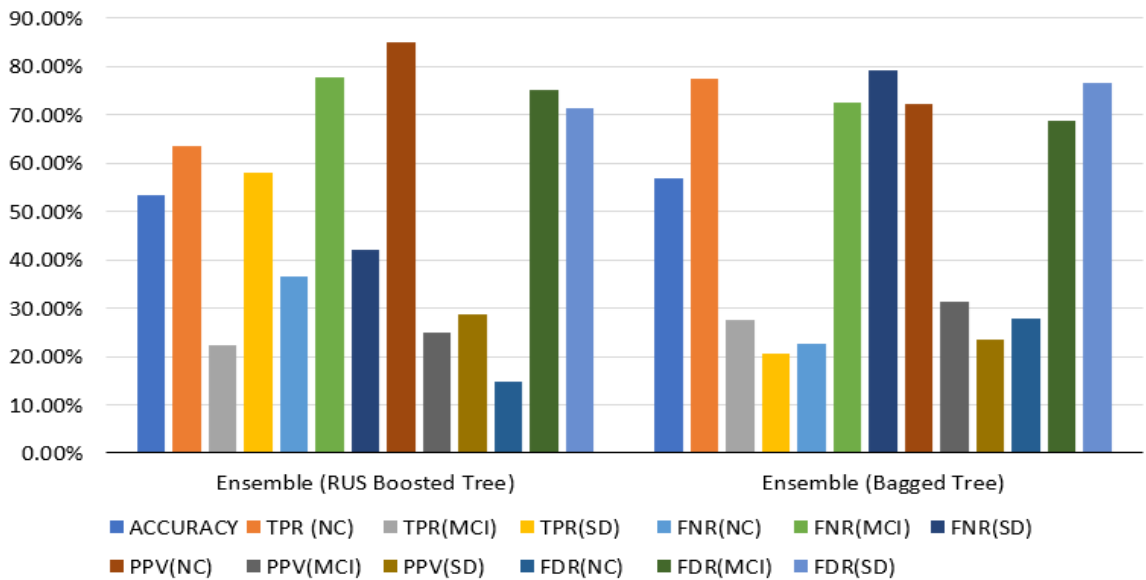

FIGURE 7. Performance comparison of multiclass classification models with MMSE subdivided patients, obtained for the imbalanced dataset.

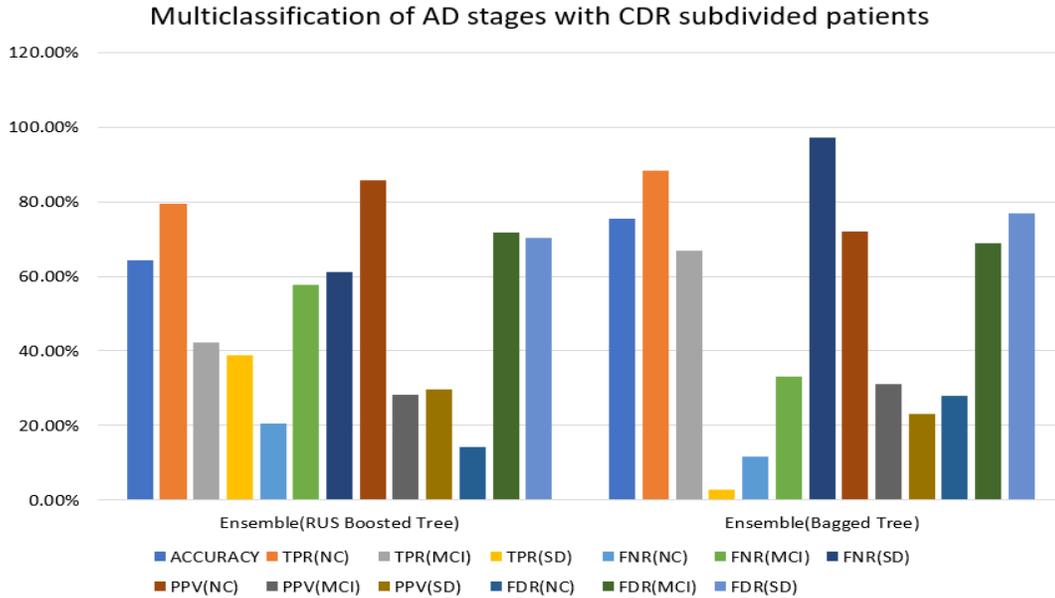

FIGURE 8. Performance comparison of multiclass classification models with CDR subdivided patients, obtained for the imbalanced dataset.

classifier had better accuracy (63.0%) in comparison to the boosted classifier. Moreover, the bagged tree classifier showed high TPR values for the SD group in both MMSE and CDR-staged patients. Hence, overall, Bagged Tree demonstrates better performance than Ensemble Boosted Tree. The confusion matrix of the Ensemble Bagged Tree model for MMSE-staged patients is shown in Figure 9a. It can be observed that 62 individuals were correctly classified as NC, 50 patients were classified as MCI, and 56 as SD. Figure 9b shows the ROC curve for the Ensemble Bagged Tree model in which Area Under the Curve (AUC) was found to be 0.71. Figure 9c shows the confusion matrix of the Ensemble Bagged tree classifier that classifies 23 individuals as NC, 22 patients as MCI, and 23 as SD for CDR-staged patients. Figure 9d shows the ROC curve for the Ensemble Bagged Tree model where AUC was found to be 0.81. Table VI and Table VII show the comparison of Ensemble Boosted and Bagged Tree models that we used for the multi-classification of MMSE and CDR-staged patients after balancing the dataset. Ensemble Bagged

Tree showed better performance for both types of patient staging. Figures 10 and 11 show the comparison of the models in terms of accuracy, TPR, FNR, PPV, and FDR. Again, the Bagged Tree showed higher accuracy as well as higher TPR and lower FNR in SD groups.

Hence, it can be concluded that balancing the dataset resulted in the same machine learning classifier (i.e., Ensemble Bagged Tree) providing higher accuracy as well as higher TPR and lower FNR in the SD group for both MMSE and CDR-staged patients. In contrast, in the imbalanced dataset, the NC class was over-represented resulting in one machine learning classifier providing higher accuracy whereas another providing higher TPR and lower FNR in the SD group.

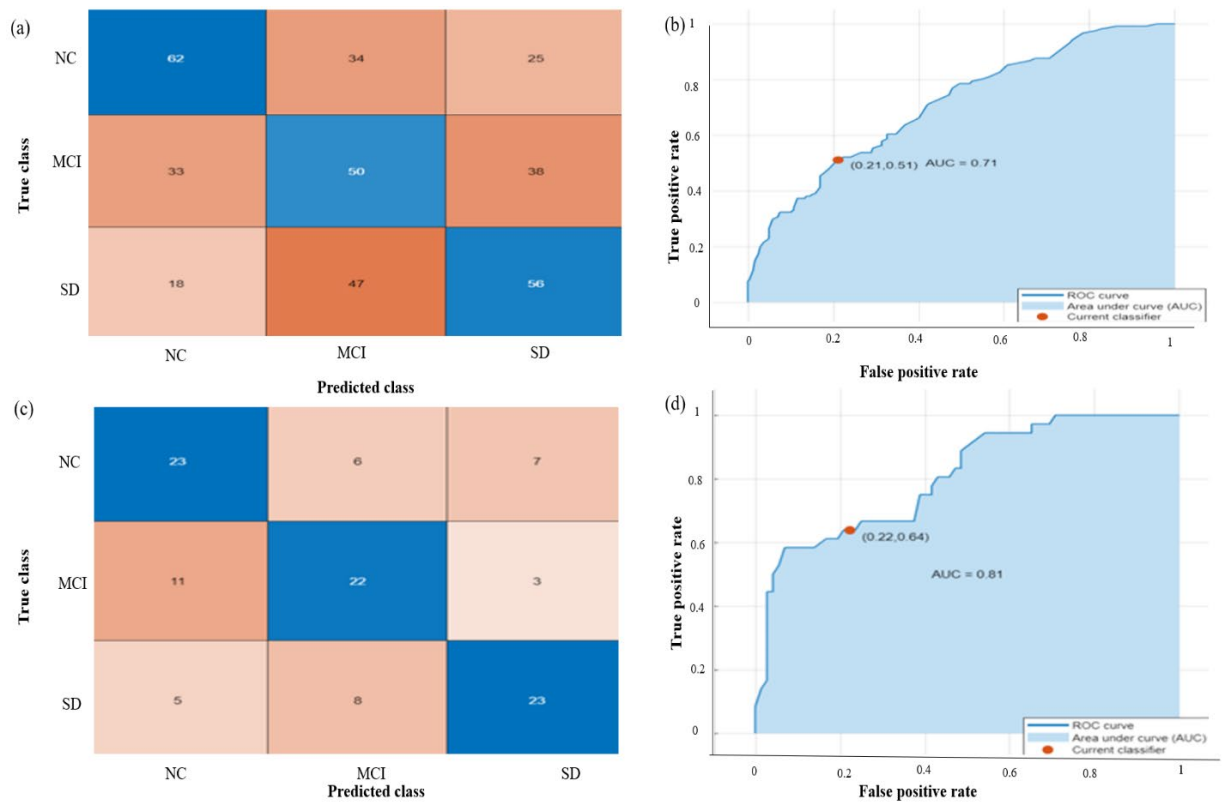

**FIGURE 9.** Machine learning classification applied to a balanced dataset. (a) Confusion matrix and (b) Receiver operating characteristic (ROC) curve for multiclass classification of MMSE subdivided patients using Ensemble Bagged Tree. (c) Confusion matrix and (d) ROC curve for multiclass classification of CDR subdivided patients using Bagged Tree.

TABLE VI
COMPARISON OF CLASSIFICATION MODELS FOR MULTI CLASSIFICATION OF MMSE-STAGED PATIENTS, OBTAINED WITH A BALANCED DATASET.

| Machine Learning Models | Accuracy | TPR (NC) | TPR (MCI) | TPR (SD) |
|---|---|---|---|---|
| Ensemble (Boosted Tree) | 44.6% | 51.2% | 35.5% | 47.1% |
| Ensemble (Bagged Tree) | 46.3% | 51.2% | 41.3% | 46.3% |

TABLE VII
COMPARISON OF CLASSIFICATION MODELS FOR MULTI CLASSIFICATION OF CDR-STAGED PATIENTS, OBTAINED WITH A BALANCED DATASET.

| Machine Learning Models | Accuracy | TPR (NC) | TPR (MCI) | TPR (SD) |
|---|---|---|---|---|
| Ensemble (Boosted Tree) | 60.2% | 61.1% | 61.1% | 58.3% |
| Ensemble (Bagged Tree) | 63.0% | 63.9% | 61.1% | 63.9% |

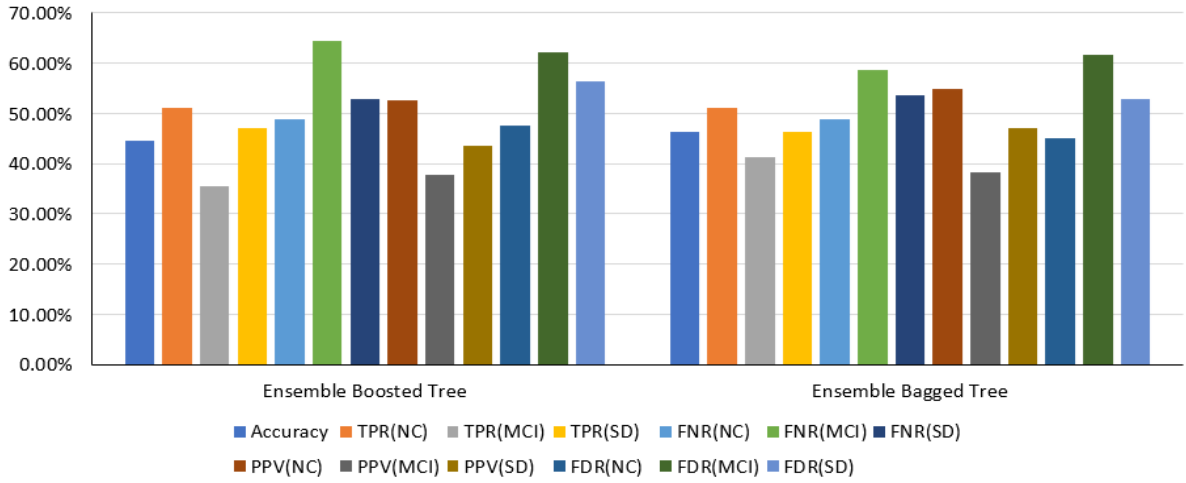

FIGURE 10. Performance comparison of multiclass classification models with MMSE subdivided patients, obtained for a balanced dataset.

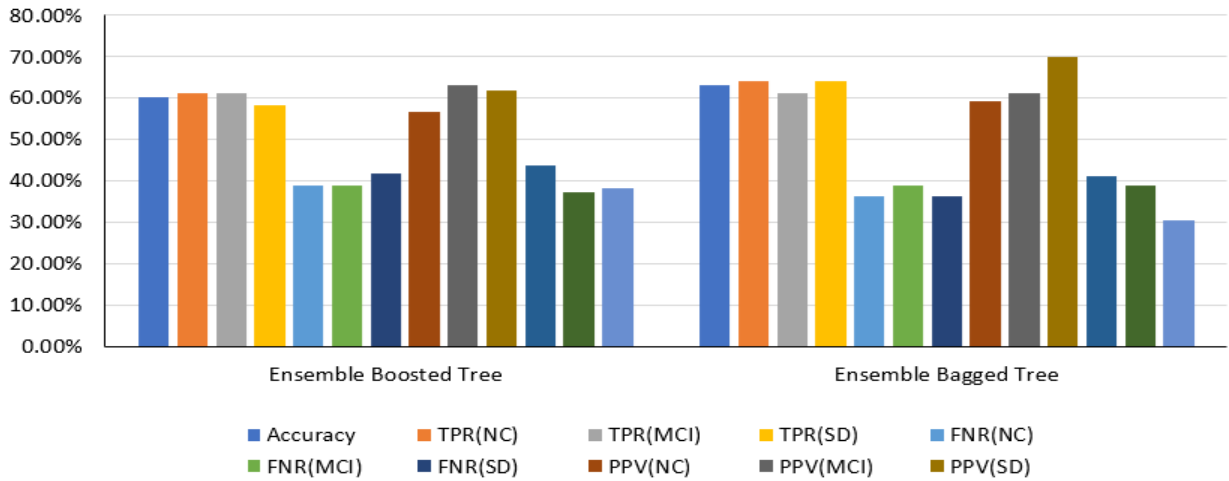

FIGURE 11. Performance comparison of multiclass classification models with CDR subdivided patients, obtained for a balanced dataset.

## Results and Discussion

This study aims to determine whether CSF biomarkers ($A\beta1\text{-}42$, T-tau, P-tau, and $A\beta1\text{-}42$/P-tau) alone can be used to classify different stages of Alzheimer's disease. The performance of the classification models was compared for CDR and MMSE-staged patients. To identify the most effective multi- (NC vs MCI vs SD) and binary classification (AD vs NC) models, we have evaluated several machine learning models using several metrics including accuracy, TPR, FNR, PPV, and FDR. Previous research also reports the classification of Alzheimer's disease stages using binary classification (NC vs. AD [14] and MCI vs. SD [15]) or multiclass classification (NC vs. MCI vs. SD) [16] models. However, the majority of the studies were regarded as subjective or culturally influenced (age, history of genetic relationships with family members, smoking, and educational background). Other studies used CSF biomarker levels along with other sophisticated techniques like magnetic resonance imaging (MRI) or positron emission tomography (PET) imaging. In contrast, this work shows the potential of CSF biomarkers alone ($A\beta1\text{-}42$, T-tau, P-tau, and $A\beta1\text{-}42$/P-tau) to distinguish between different stages of AD. Additionally, we have demonstrated that the CDR scoring outperforms the MMSE scoring in the classification of AD stages for both binary and multi-classifications. For CDR scoring-based binary classification, the accuracy of the Ensemble Boosted Tree classifier was found to be 84.4%, whereas, for MMSE scoring, the accuracy of the Ensemble Boosted Tree classifier was 73.3%. On the other hand, the RUS boosted tree classifier showed an accuracy of 64.2% for CDR- subdivided patients, while the boosted tree classifier had an accuracy of 53.4% for MMSE-subdivided patients, both for the multiclassification study with an imbalanced dataset. With the balanced dataset, Ensemble Bagged Tree provided higher accuracy as well as higher TPR and lower FNR in the SD group for both MMSE and CDR-staged patients.

Additionally, results showed that for binary classification using MMSE-subdivided patients, the accuracy, TPR, and FNR values of Ensemble Boosted Tree outperformed other machine learning classifiers. For binary classification using CDR-subdivided patients, Ensemble Boosted Tree and Bagged Tree outperformed other machine learning classifiers in terms of accuracy, TPR, and FNR.

In summary, CSF biomarkers offer potential in the early detection of AD. This study tested the hypothesis that distinct stages of AD are correlated with different levels of CSF biomarkers. The results are encouraging from a future standpoint because they show good accuracy with fewer input features. The model's accuracy could be improved by including more patient data and real-time input features. There is evidence in the literature that a strong correlation exists between blood biomarkers and CSF biomarkers [34]. Hence, future research could also involve developing a point-of-care blood sampling device interfaced with machine-learning classification models for early diagnosis and prognosis of Alzheimer's disease patients.

## Appendix

In this work, we used the following classification models: -

*Ensemble Learning:* - By utilizing ensemble algorithms, a collection of "weak" learners is combined to form a higher-performing ensemble model [35, 36]. Boosting, a powerful approach initially developed for classification in the field of machine learning [37], is considered as one of the main techniques for creating such ensembles. Both bagging and boosting are popular approaches for building ensembles, involving the utilization of fundamental learners [36]. The evaluation of boosted trees includes the AdaBoost method in combination with decision trees [35]. AdaBoost, a boosting method, is commonly employed in ensemble models to improve the classification performance of weak learners while minimizing the risk of overfitting [38]. Another technique utilized in ensemble learning is the bagged tree, which combines the bagging technique with decision trees [35].

*Naïve Bayes classifier:* - Naïve Bayes Classifiers are based on the fundamental principle of Bayes Theorem and rely on the assumption of conditional independence among attributes. These classifiers operate under the assumption that the presence of one feature has no correlation with the presence of any other feature. Naïve Bayes Classifiers are particularly effective in supervised learning settings. They are capable of estimating classification parameters with a small amount of training data. These classifiers are straightforward to build and utilize, making them suitable for a wide range of real-life applications.

*Support Vector Machine (SVM):* - SVM was designed to classify datasets with distinct categories. Its classification process involves finding the optimal hyperplane that maximizes the margin between these classes. In the case of linear SVM, data is divided using a straight line. However, linear separation is often insufficient for SVM to effectively classify data. As a result, SVM necessitates a more intricate structural approach to achieve optimal separation. The introduction of kernel functions addresses this limitation. By employing kernel functions, data is mapped into a higher dimensional space, allowing SVM to handle nonlinearly separable classes and achieve linear separability in this transformed space.

*K-Nearest Neighbors (KNN): -* The k-nearest neighbors' algorithm, commonly known as KNN or k-NN, is a supervised learning classifier that utilizes proximity to classify or predict the category of an individual data point.

*Logistic Regression: -* Logistic regression is a method used to learn functions of the form f: X → Y or P(Y|X), where Y represents discrete values and X is a vector containing discrete or continuous variables. In logistic regression, the distribution P(Y|X) is assumed to have a parameterized shape, and these parameters are directly estimated from the training dataset. This approach bypasses the two-step calculation method of Naïve Bayes in estimating P(Y|X). Logistic regression is often referred to as a discriminative classifier because it views the distribution P(Y|X) as directly determining the target value Y for a given instance X. According to [39], when the target variable C is Boolean and the assumptions of Gaussian Naïve Bayes (GNB) hold true, the Gaussian Naïve Bayes (GNB) and logistic regression classifiers asymptotically converge towards identical results as the number of training examples approaches infinity. However, it's worth noting that GNB parameter estimates converge to their asymptotic values with a logarithmic relationship to the number of examples (log n), where n represents the dimension of X. On the other hand, logistic regression parameter estimates require a linear relationship with the number of examples (order n) and converge at a slower rate [38].

*True Positive Rate (TPR): -* TPR, also known as sensitivity or recall, represents the proportion of accurate predictions made for the positive class. TPR is calculated using the following equation (1):

$$TPR = \frac{True\ Positive}{True\ Positive + False\ Negative} \qquad (1)$$

True positive is an outcome where the model correctly predicts the positive class. Similar to a true positive, a true negative is an outcome in which the model accurately predicts the negative class.

*False Negative Rate (FNR):* - FNR is the proportion of positive instances that are incorrectly classified as negative by the test. FNR is calculated by equation (2):

$$FNR = \frac{False\ Negative}{False\ Negative + True\ Positive} \qquad (2)$$

False negative is the outcome where the model incorrectly predicts the negative class.

*Positive predictive value (PVV):* - The number of correctly classified positive samples divided by the total number of samples that were classified as positive.

*False Discovery Rate (FDR):* - A statistical measure that quantifies the proportion of falsely rejected null hypotheses among all rejected hypotheses.

## Acknowledgement:

We collected data from the NACC database. The NACC database is funded by NIA/NIH Grant U01 AG016976. NACC data are contributed by the NIA-funded ADRCs: P30 AG019610, P30 AG013846, P50 AG008702, P50 AG025688, P50 AG047266, P30 AG010133, P50 AG005146, P50 AG005134, P50 AG016574, P50 P30 AG053760, P30 AG010124, P50 AG005133, P50